\documentclass{article} 
\usepackage{iclr2027_conference,times}

\usepackage{amsmath,amsfonts,bm}

\def\eqref#1{equation~\ref{#1}}

\def\1{\bm{1}}

\DeclareMathAlphabet{\mathsfit}{\encodingdefault}{\sfdefault}{m}{sl}
\SetMathAlphabet{\mathsfit}{bold}{\encodingdefault}{\sfdefault}{bx}{n}

\usepackage{hyperref}
\usepackage{url}
\usepackage{graphicx}
\usepackage{booktabs}
\usepackage{multirow}

\usepackage{enumitem}
\usepackage{float}
\usepackage{wrapfig}

\usepackage{amsmath,amssymb, bm,booktabs}
\usepackage{algorithm}
\usepackage{algpseudocode}

\usepackage{fancyhdr}
\fancypagestyle{labpaperfirst}{%
  \fancyhead{}
  \fancyhead[L]{\includegraphics[width=4.2in]{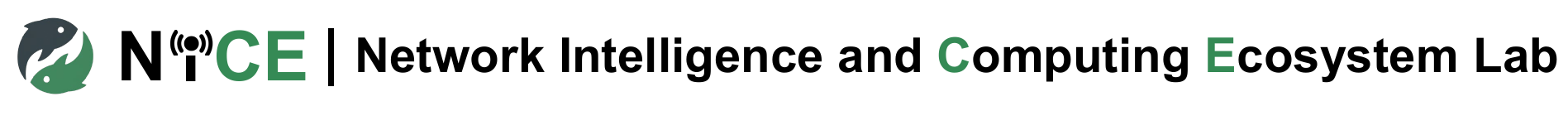}}%
  \renewcommand{\headrulewidth}{0pt}%
}

\title{Carnator: Fast Text-to-Video Generation with Generation-Native Compatibility-Guided Cross-Request Reuse}

\author{
Xingkun Yin$^{1}$ \quad
Xuebin Tang$^{1}$ \quad
Mingkun Xu$^{2}$ \quad
Hongyang Du$^{1,\dagger}$ \\
$^1$Department of Electrical and Computer Engineering, \\ University of Hong Kong, Hong Kong SAR, China \\
$^2$Guangdong Institute of Intelligence Science and Technology, Zhuhai, China \\
$^\dagger$Corresponding authors. \\
\texttt{yinxingkun@connect.hku.hk, marcus.tang.official@gmail.com} \\
\texttt{xumingkun@gdiist.cn, duhy@hku.hk}
}

\iclrfinalcopy 
\begin{document}

\maketitle
\thispagestyle{labpaperfirst}   
\lhead{Preprint}                
\renewcommand{\headrulewidth}{0.4pt}

\begin{abstract}
Video diffusion transformers produce high-quality videos, yet iterative denoising incurs substantial inference latency, limiting interactive and large-scale serving.
Most existing acceleration methods focus on individual requests, thereby restricting efficiency gains to redundancy within a single generation trajectory.
Recent cross-request reuse offers an additional source of savings, but existing approaches often infer reusability from coarse semantic similarity. 
This conflates semantic relatedness with generation-level computational compatibility, so aggressive reuse may accept incompatible historical computation while conservative reuse leaves substantial acceleration unrealized.
We present \emph{Carnator}, a cross-request acceleration framework that addresses this challenge by extracting and using generation-native compatibility evidence directly from the model's evolving internal states.
Specifically, \emph{Carnator} performs a lightweight early probe to construct an Early Signature from internal diffusion states, assessing reuse validity through risk-aware compatibility decisions.
The same evidence characterizes reuse scope by localizing target-specific computation and guiding joint reuse of historical latent trajectories and sparse attention connectivity.
Across three text-to-video backbones, Carnator consistently achieves higher cache-hit end-to-end acceleration than the evaluated cross-request baselines despite more selective cache acceptance, reaching up to 2.17$\times$ speedup while maintaining competitive generation quality.
\end{abstract}

\section{Introduction}
Video diffusion transformers have rapidly advanced in generation fidelity, temporal coherence, and controllability, fueling the growing popularity and adoption of AI-generated video across content creation, entertainment, and online media~\citep{wan2025wan,kong2024hunyuanvideo,yang2025cogvideox}.
However, video generation remains computationally intensive as iterative denoising repeatedly processes long spatiotemporal token sequences through large Transformer backbones.
This results in substantial generation latency and prolonged GPU occupation, limiting both interactive use and serving throughput.
For instance, in batch-one video generation, where lower latency improves throughput, a 3.2$\times$ speedup can reduce serving cost by about 69$\%$ at scale, highlighting the practical value of efficient inference \citep{shahsavan2026wan22}.

To accelerate video generation, prior work has primarily explored a broad range of intra-request acceleration techniques.
These approaches exploit computational redundancy, including few-step distillation, feature caching~\citep{Nie_2026_CVPR, lyu2025fastercache}, token-wise computation reduction, and sparse attention~\citep{ICLR2025_bbe024e0, chen2026sparse}.
While effective, their acceleration gains are inherently bounded by the redundancy available within a single request. 
At serving scale, however, video generation systems operate under continuous workloads consisting of requests that often share objects, scenes, and structural patterns.
This workload pattern reveals a new opportunity for inference acceleration by exploiting computational redundancy across different generation requests. 

\begin{wrapfigure}{R}{0.52\linewidth}
\vspace{-8pt}
\centering
\includegraphics[width=\linewidth]{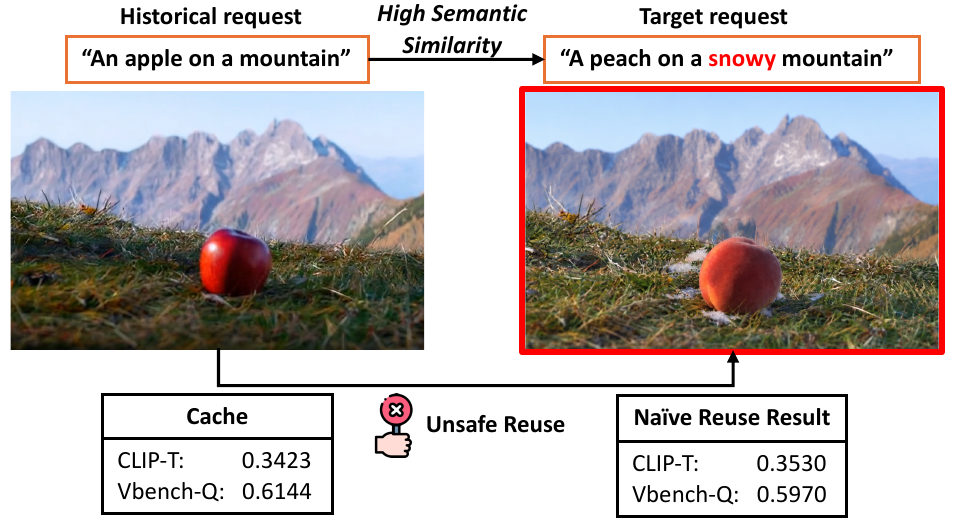}
\caption{
Failure of semantic-similarity-based reuse. A semantically similar historical request yields competitive CLIP-T and VBench-Q scores, yet naive reuse preserves the incompatible non-snowy background and violates the target request. Carnator detects the incompatibility and falls back to dense generation.
}
\label{fig:motivation}
\vspace{-10pt}
\end{wrapfigure}

Recent cross-request methods have demonstrated that historical computation can accelerate related generation requests.
However, effective reuse requires two coupled decisions, \emph{reuse validity}, which determines whether a historical cache can be reused, and \emph{reuse scope}, which determines where target-specific computation must be preserved once reuse is deemed valid.
Reuse validity cannot be reliably inferred from request semantics alone, as semantically similar requests may develop substantially different visual structures during generation.
As illustrated in Fig.~\ref{fig:motivation}, an incompatible reuse can preserve historical scene content that conflicts with the target request while still achieving competitive aggregate quality scores.
Existing text-to-video (T2V) approaches have primarily exploited intermediate latent states or region-selective recomputation, while recent sparse-attention connectivity reuse has been demonstrated mainly in image-to-video (I2V) settings with stronger visual conditioning.
Reuse scope therefore remains unclear in T2V, since reusable computation may span both latent trajectories and attention connectivity.
Even for a compatible history, reusable computation may vary across spatial regions and attention interactions.
Overly aggressive reuse may degrade generation quality, whereas conservative reuse leaves substantial acceleration opportunities unrealized.

These challenges suggest looking beyond request semantics and into the generation process itself.
Early diffusion states already reveal request-specific differences in object appearance, geometry, and spatial structure~\citep{wen2024detecting,liu2024towards,qi2023fatezero}.
Cross-attention provides noun-conditioned localization, self-attention expands this evidence over the spatial support of generated objects, and the corresponding latent states encode emerging geometric and appearance information.
We refer to these request-specific internal signals as \emph{generation-native compatibility evidence}, which better captures reuse validity than prompt semantics alone by directly reflecting the evolving target generation.
The same evidence also reveals object regions and attention interaction structure, indicating where target-specific computation must be preserved and where historical computation can remain reusable.
Together, these observations suggest that generation-native compatibility evidence can support both reuse validity and reuse scope within a unified cross-request acceleration framework.

Motivated by these observations, we present \emph{Carnator}, a generation-native compatibility-guided framework for cross-request video generation acceleration.
To assess reuse validity, \emph{Carnator} first performs text-level recall to retrieve semantically related historical requests and then applies a lightweight early probe to construct an Early Signature, a compact representation of generation-native compatibility evidence in the target's early generation state.
The Early Signature is compared with recalled histories under risk-aware compatibility thresholds to select caches whose computation remains suitable for reuse.
Once a cache is accepted, the same evidence is carried forward to determine reuse scope by identifying the target-specific computation that must be preserved.
For accepted caches, it guides \emph{Joint Latent-Attention Reuse}, which combines reuse of historical latent states and sparse attention connectivity while preserving target-specific computation.
Extensive experiments across multiple video diffusion models demonstrate that \emph{Carnator} consistently reduces end-to-end (E2E) inference latency while maintaining competitive generation quality.

Our main contributions are summarized as follows:
\begin{itemize}[leftmargin=*]
    \item \emph{Carnator} introduces a generation-native compatibility assessment component for reuse validity. 
    It performs a lightweight early probe to construct an Early Signature directly from internal diffusion states and uses risk-aware compatibility decisions to identify compatible historical computation beyond coarse semantic similarity.

    \item \emph{Carnator} further develops an evidence-guided joint reuse component for reuse scope. The same generation-native evidence localizes target-specific computation and guides joint reuse of historical latent trajectories and sparse attention connectivity, reducing redundant token updates and attention interactions while preserving necessary target updates.


    \item We conduct extensive experiments on VidProM across three text-to-video diffusion backbones to evaluate both reuse validity and inference acceleration. \emph{Carnator} improves cache selection reliability, reduces unsafe reuse, and achieves up to 1.72$\times$ end-to-end acceleration across the three backbones while maintaining competitive generation quality.
\end{itemize}

\section{Related Work}

\paragraph{Efficient diffusion inference.}
Diffusion models require repeated neural network evaluation over many denoising steps, and video generation further increases the cost through long spatiotemporal token sequences. Existing acceleration methods reduce this cost through step reduction, quantization, token selection, feature caching, sparse attention, and adaptive computation. Recent approaches also coordinate multiple acceleration mechanisms to reduce computation across timesteps, tokens, and attention connections \citep{ICLR2026_921ac785,zhao2026rapid}. Most of these techniques optimize a single generation process and do not exploit reusable computation from previously completed requests.

\paragraph{Intra-request acceleration.}
Intra-request methods exploit redundancy within one denoising trajectory. Feature caching reuses intermediate representations across nearby timesteps, while token-wise methods selectively refresh tokens according to their sensitivity to caching errors \citep{ICLR2025_bbe024e0,liu2025speca}. Sparse attention methods reduce quadratic attention cost by identifying structured patterns across timesteps, layers, and heads, then executing only selected query-key interactions with efficient kernels \citep{chen2026sparse}. Diffusion attention has also been used to recover semantic regions from cross-attention and self-attention maps \citep{tian2024diffuse}. These methods provide important foundations for selective computation, but their reuse decisions are derived from states within the current trajectory. \emph{Carnator} instead considers whether latent states, intermediate features, token regions, and attention patterns from a different generation trajectory remain compatible with the target request.

\paragraph{Inter-request acceleration.}
Inter-request reuse has recently emerged as a new direction for diffusion serving. Early work retrieves similar T2I requests and reuses intermediate noise states to skip initial denoising computation \citep{agarwal2024approximate}. Subsequent methods extend this idea to T2V generation through entity-level retrieval, shared semantic components, staged latent reuse, and region-selective recomputation \citep{liu2026beyond}. 
CHAI~\citep{cherian2026chai} instantiates cross-inference cache attention on OpenSora 1.2 by modifying its STDiT attention blocks, while its portability to other video diffusion architectures remains unexplored.
Historical sparse attention patterns can also be transferred across related I2V requests to reduce online mask construction \citep{liu2026chorus}. 
\emph{Carnator} uses generation-native compatibility evidence to address reuse validity and reuse scope in inter-request acceleration.
It constructs an Early Signature for compatibility filtering according to the target generation state, and exploits accepted caches through joint region-guided latent reuse and historical sparse-connectivity reuse.

\section{Method}

\subsection{Overview and Problem Setup}

\begin{figure*}[t]
    \centering
    \includegraphics[width=\textwidth]{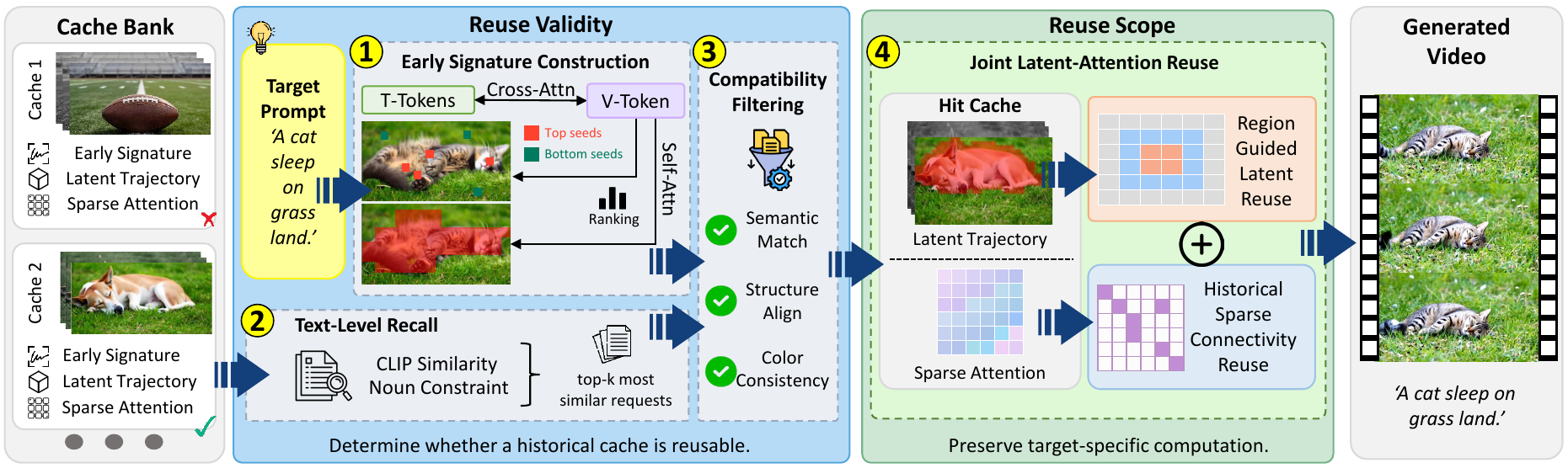}
    \caption{
    Overview of \emph{Carnator}.
    Historical requests are first retrieved through text-level recall and filtered using generation-native Early Signatures.
    For an accepted cache, \emph{Carnator} jointly reuses historical latent trajectories and sparse attention connectivity while preserving target-specific computation.
    }
    \label{fig:overview}
\end{figure*}

Let $q$ denote a target video generation request and $\mathcal{B}=\{\mathcal{C}_1,\ldots,\mathcal{C}_M\}$ a cache bank constructed from completed historical requests.
For each target request, \emph{Carnator} addresses two coupled decisions, namely reuse validity, which determines whether a historical cache can be reused, and reuse scope, which determines where target-specific computation must be preserved within joint latent-attention reuse once a compatible cache is identified.

To assess reuse validity, \emph{Carnator} first performs text-level recall to identify semantically related candidate histories and then runs a lightweight target probe to construct an Early Signature $S_q$, a compact representation of generation-native compatibility evidence.
The target signature is compared with the stored signatures of recalled caches under risk-aware compatibility thresholds. If no candidate is compatible, the target proceeds with dense generation.
Otherwise, an accepted cache $\mathcal{C}_{j^\star}$ is selected for subsequent reuse.

For an accepted cache, \emph{Carnator} carries the same generation-native compatibility evidence forward to determine reuse scope.
The localized evidence identifies where target-specific computation must be preserved within joint latent-attention reuse.
The same generation-native evidence therefore connects the two stages, with the Early Signature supporting reuse-validity assessment and its localized evidence guiding reuse scope during subsequent inference.
The overall workflow is illustrated in Fig.~\ref{fig:overview} with more details in Appendix~\ref{app:early-signature}. 
Sec.~\ref{subsec-cache_compatibility} describes reuse validity, while Sec.~\ref{subsec-joint_reuse} describes reuse scope.

\subsection{Reuse Validity with Generation-Native Evidence}
\label{subsec-cache_compatibility}

\paragraph{Early Signature Construction}
\emph{Carnator} requires generation-native evidence before committing to cache reuse.
For a request $r$ with prompt $y_r$, it performs a lightweight early probe and summarizes the partially formed generation state into an Early Signature $S_r$.
Conceptually, $S_r$ is a compact representation of generation-native evidence that is informative for cross-request compatibility.
In this work, we instantiate $S_r$ using lightweight object-centric spatiotemporal support and appearance descriptors extracted from the model's internal cross- and self-attention states.
Early diffusion states already provide informative evidence for reuse-validity assessment, as illustrated in Fig.~\ref{fig:early_signature_evidence}, with additional implementation details provided in Appendix~\ref{app:early-signature}.

\begin{figure*}[t]
    \centering
    \includegraphics[width=\textwidth]{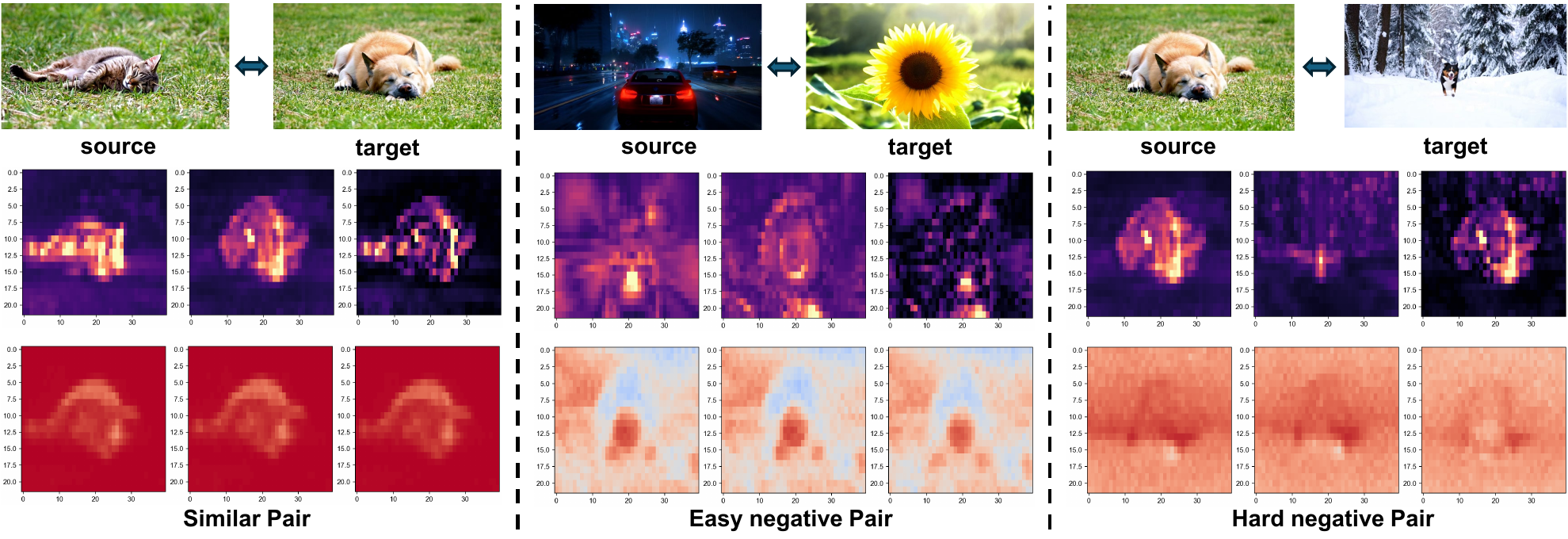}
    \caption{
    Early generation states provide informative signals for cross-request compatibility.
    Similar request pairs exhibit stronger alignment in early denoising representations, while easy and hard negative pairs show increasingly distinct prompt-conditioned residuals and latent dynamics.
    }
    \label{fig:early_signature_evidence}
\end{figure*}

For each noun entity $o\in\mathcal{N}(y_r)$, \emph{Carnator} first extracts noun-conditioned seed responses from cross-attention, following prior use of cross- and self-attention for semantic localization~\citep{liu2024towards,qi2023fatezero}.
Let $\Omega$ denote the spatiotemporal visual-token domain over $T\times H\times W$ locations. 
For visual token $i\in\Omega$, $A^{\mathrm{cross}}_{l,h}[i,o]$ denotes its cross-attention weight to noun $o$ at Transformer layer $l$ and head $h$.

These responses are aggregated as
$C_i^o=\mathrm{Agg}_{l,h}A^{\mathrm{cross}}_{l,h}[i,o]$,
with high- and low-response tokens forming the noun and noise seed sets $\mathcal{I}_o^{+}$ and $\mathcal{I}_o^{-}$, respectively.
The seeds are then expanded through self-attention using
$
S_i^o=
\operatorname{Agg}_{h}
\left[
\max_{u\in\mathcal I_o^{+}}A_h^{\mathrm{self}}[i,u]
-
\max_{u\in\mathcal I_o^{-}}A_h^{\mathrm{self}}[i,u]
\right].
$
Fusing $C_i^o$ and $S_i^o$ and thresholding the resulting object-support score yields the Evidence Region
$R_{r,o}^{\mathrm{evidence}}\subseteq\Omega$.

\emph{Carnator} converts the Evidence Region into a binary spatiotemporal indicator
$m_{r,o}\in\{0,1\}^{T\times H\times W}$, which summarizes the object's support across both video frames and spatial locations.
The compatibility descriptors are read from the tail slice
$m_{r,o}(T{-}1,\cdot,\cdot)$, since appearance is recovered from a single decoded frame.
Its normalized support size is therefore
$$
A_{r,o}
=
\frac{1}{HW}
\sum_{h,w} m_{r,o}(T{-}1,h,w).
$$
Appearance is represented by the mean CIELAB color
$\mathbf{c}_{r,o}^{\mathrm{Lab}}\in\mathbb{R}^3$
over the corresponding decoded tail frame.
The resulting descriptor is
$
s_{r,o}
=
(m_{r,o},A_{r,o},\mathbf{c}_{r,o}^{\mathrm{Lab}}).
$

Under this object-centric instantiation, we construct the Early Signature as
$
S_r=\{s_{r,o}\mid o\in\mathcal{N}(y_r)\}.
$
The same representation can accommodate additional generation-native statistics when compatibility is characterized by richer temporal or inter-object structure.
For historical requests, the resulting signature is stored together with the reusable latent states and sparse connectivity.

\paragraph{Compatibility Filtering}
Given a target request $q$ and cache bank $\mathcal{B}$, \emph{Carnator} first performs text-level recall to obtain a coarse candidate set $\mathcal{L}_q$.
The target constructs its Early Signature $S_q$ once and compares it with the stored signature $S_j$ of each candidate $\mathcal{C}_j$.
Compatibility is evaluated over the entity correspondences established during candidate recall.

The generation-native compatibility is represented by a
discrepancy vector
$
\mathbf{d}(q,\mathcal{C}_j)
=
\left[d_k(q,\mathcal{C}_j)\right]_{k\in\mathcal{D}(q,\mathcal{C}_j)},
$
where each $d_k$ measures a compatibility-relevant discrepancy between the
target and historical generation states, and
$\mathcal{D}(q,\mathcal{C}_j)$ denotes the set of applicable discrepancy
dimensions for the candidate pair.
The resulting discrepancy vector captures four complementary aspects of generation-native compatibility, including context appearance, object size, object shape, and object appearance.
Shared entities contribute the context-appearance discrepancy.
When a substituted entity pair is present, \emph{Carnator} additionally compares its size, object shape, and appearance.
For the shared nouns, \emph{Carnator} measures the maximum appearance discrepancy
$
d_{\mathrm{context}}
=
\max_{o\in\mathcal{O}_{\mathrm{same}}}
\Delta E_{00}
\left(
\mathbf{c}^{\mathrm{Lab}}_{q,o},
\mathbf{c}^{\mathrm{Lab}}_{j,o}
\right),
$
where $d_{\mathrm{context}}=0$ when $\mathcal{O}_{\mathrm{same}}$ is empty.
When a substituted entity pair is present, \emph{Carnator} further compares its size, object shape, and appearance as
\[
d_{\mathrm{size}}
=
\left|
\log\frac{A_{q,o_q}}{A_{j,o_j}}
\right|,
\qquad
d_{\mathrm{shape}}
=
1-
\frac{
2|\widetilde{m}_{q,o_q}\cap\widetilde{m}_{j,o_j}|
}{
|\widetilde{m}_{q,o_q}|+|\widetilde{m}_{j,o_j}|
},
\qquad
d_{\mathrm{color}}
=
\Delta E_{00}
\left(
\mathbf{c}^{\mathrm{Lab}}_{q,o_q},
\mathbf{c}^{\mathrm{Lab}}_{j,o_j}
\right).
\]
Here, $\widetilde{m}_{q,o_q}$ and $\widetilde{m}_{j,o_j}$ denote bounding-box-normalized object indicators, obtained by cropping each tail slice to its tight support box and resizing it to a common 2D resolution.

Given a threshold $\tau_k$ for each applicable discrepancy dimension,
the compatibility decision is
$$
g_{\boldsymbol{\tau}}(q,\mathcal{C}_j)
=
\prod_{k\in\mathcal{D}(q,\mathcal{C}_j)}
\mathbb{I}
\left[
d_k(q,\mathcal{C}_j)\le\tau_k
\right].
$$
\emph{Carnator} selects the first compatible candidate in recall order and falls back to dense generation if none satisfies the rule.

\paragraph{Risk-Aware Compatibility Decisions.}
\emph{Carnator} selects compatibility thresholds $\boldsymbol{\tau}$ from
observed reuse outcomes.
For each request--cache pair $i$, let $Q_i^{(m)}$ and
$Q_i^{\mathrm{dense},(m)}$ denote the quality of the reused and dense
generations for the same target request under criterion $m\in\mathcal{M}$.
We define an acceptable reuse outcome as
\[
y_i =
\mathbb{I}
\left[
Q_i^{(m)}
\ge
Q_i^{\mathrm{dense},(m)}-\Delta_m,
\ \forall m\in\mathcal{M}
\right],
\]
where $\Delta_m$ specifies the tolerated reuse-induced degradation under
criterion $m$.
This relative criterion normalizes prompt-dependent generation difficulty and
isolates quality degradation introduced by reuse.

On the threshold-selection set, \emph{Carnator} selects
$\boldsymbol{\tau}$ to maximize reuse coverage subject to an unsafe-reuse
risk constraint,
$
\boldsymbol{\tau}^{\star}
=
\arg\max_{\boldsymbol{\tau}} C(\boldsymbol{\tau})
\quad
\text{s.t.}
\quad
R(\boldsymbol{\tau})\le\delta,
$
with
\[
C(\boldsymbol{\tau})
=
\frac{1}{N}\sum_i g_{\boldsymbol{\tau}}(i),
\qquad
R(\boldsymbol{\tau})
=
\frac{\sum_i g_{\boldsymbol{\tau}}(i)(1-y_i)}
{\sum_i g_{\boldsymbol{\tau}}(i)}.
\]
The selected thresholds are then fixed and evaluated on a disjoint
certification set. For $n$ accepted pairs with $u$ unsafe cases, we compute
the one-sided Clopper--Pearson upper confidence bound $U_\alpha(u,n)$ and
accept the thresholds when $U_\alpha(u,n)\le\delta$.

\subsection{Reuse Scope with Joint Latent-Attention Reuse}
\label{subsec-joint_reuse}

\begin{wrapfigure}{R}{0.48\linewidth}
\vspace{-8pt}
\centering
\includegraphics[width=\linewidth]{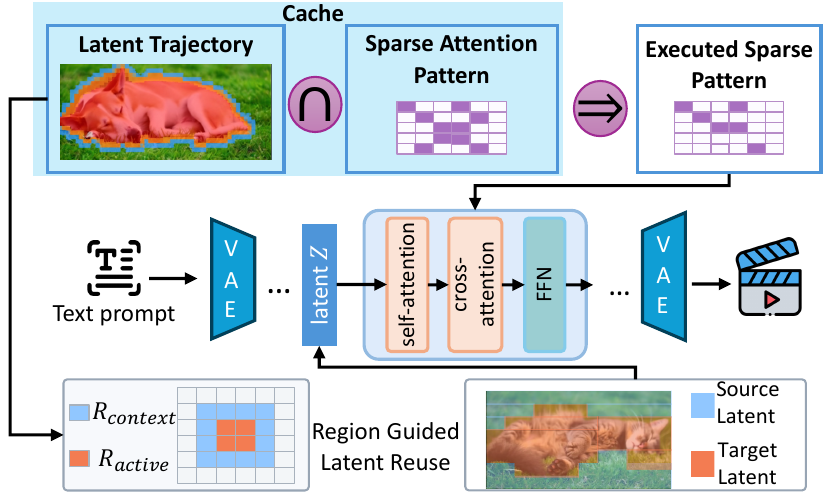}
\caption{Joint latent-attention reuse in \emph{Carnator}.}
\label{fig:joint_reuse}
\vspace{-10pt}
\end{wrapfigure}

Reuse validity determines whether a historical cache can be used.
Once accepted, \emph{Carnator} uses the same generation-native compatibility evidence to determine where target-specific computation must be preserved within a joint latent-attention reuse path.
\emph{Carnator} jointly exploits cached latent states and historical sparse attention connectivity, while restricting target-specific computation to the evidence-derived regions and their associated attention domain.
The target generation resumes from a cached intermediate state and continues denoising under the target prompt, preserving compatible historical latent states outside target-specific regions while reusing historical sparse connectivity to reduce redundant query--key interactions.
Fig.~\ref{fig:joint_reuse} illustrates how the two forms of reuse are integrated across the DiT computation path.

\paragraph{Region-Guided Latent Reuse}
For latent reuse, \emph{Carnator} uses the object regions obtained from the accepted cache to localize target-specific computation. 
The Evidence Region $\mathcal{R}_{\mathrm{evidence}}$ is spatially expanded into an Active Region $\mathcal{R}_{\mathrm{active}}$ and a larger Context Region $\mathcal{R}_{\mathrm{context}}$, satisfying
$
\mathcal{R}_{\mathrm{evidence}}
\subseteq
\mathcal{R}_{\mathrm{active}}
\subseteq
\mathcal{R}_{\mathrm{context}}.
$
The difference
$\mathcal{R}_{\mathrm{trans}}
=
\mathcal{R}_{\mathrm{active}}
\setminus
\mathcal{R}_{\mathrm{evidence}}$
forms a narrow transition band for latent blending.
All three regions are defined over the same $T\times H\times W$ spatiotemporal token domain.
Tokens in $\mathcal{R}_{\mathrm{evidence}}$ are fully updated according to the target request, tokens in the transition band receive a distance-weighted mixture of target and historical states, and tokens outside $\mathcal{R}_{\mathrm{active}}$ inherit the historical trajectory. Within selected middle Transformer layers, active tokens are recomputed using contextual keys and values from $\mathcal{R}_{\mathrm{context}}$, whereas tokens outside $\mathcal{R}_{\mathrm{active}}$ preserve their cached computation.
We focus region-restricted updates on intermediate layers, motivated by prior observations that intermediate diffusion representations contain richer visual-semantic information \citep{pnvr2023ld}, with the layer-window choice evaluated in Appendix~\ref{sec:app-layer-window}.

For each subsequent denoising step, the resulting latent is
$$
\mathbf{z}_{k+1}
=
\mathbf{W}
\odot
\widehat{\mathbf{z}}^{q}_{k+1}
+
\left(
1-\mathbf{W}
\right)
\odot
\mathbf{z}^{j^\star}_{k+1},
\label{eq_latent_reuse}
$$
where $\widehat{\mathbf{z}}^{q}_{k+1}$ denotes the target-updated latent and $\mathbf{z}^{j^\star}_{k+1}$ denotes the corresponding historical latent. The weight $\mathbf{W}\in[0,1]$ is continuous across the transition band,
$$
\mathbf{W}(x)=
\begin{cases}
1, & x\in\mathcal{R}_{\mathrm{evidence}},\\
1-s\!\left(\hat d(x)\right), & x\in\mathcal{R}_{\mathrm{trans}},\\
0, & x\notin\mathcal{R}_{\mathrm{active}},
\end{cases}
$$
with $s(\cdot)$ the smoothstep map and $\hat d(x)$ the Chebyshev distance from $\mathcal{R}_{\mathrm{evidence}}$, normalized by the maximum distance attained in the band.



\paragraph{Historical Sparse-Connectivity Reuse}
In parallel, \emph{Carnator} reuses the sparse attention structure recorded during the historical generation. For each denoising step, Transformer layer, and attention head, the historical request stores the blockwise connectivity that captures its dominant query--key interactions. During target inference, \emph{Carnator} recomputes the target $\mathbf{Q}$, $\mathbf{K}$, and $\mathbf{V}$ representations but reuses the historical sparse block structure to determine which query--key interactions are evaluated. Thus, the transferred computation describes \emph{where} attention is performed rather than reusing historical attention values themselves. This allows the target request to preserve its own representations while avoiding the dense attention computation associated with structurally redundant interactions.

\paragraph{Joint Reuse in Intermediate Layers}
The two forms of reuse are combined in the middle Transformer layers.
In these layers, historical sparse connectivity is further constrained by the target-specific computation domain defined by $\mathcal{R}_{\mathrm{active}}$ and $\mathcal{R}_{\mathrm{context}}$.
Only historical connectivity associated with active queries and their valid contextual tokens is retained, so attention reuse is restricted to the portion of the historical structure that remains relevant to the target generation. Other Transformer layers directly reuse the historical sparse block structure without this additional region constraint. 
In this way, \emph{Carnator} combines region-guided latent reuse with historical sparse-connectivity reuse, reducing computation along token and attention dimensions while preserving target-specific semantic updates.
Additional implementation details are provided in Appendix~\ref{app:early-signature}.

\section{Experiments}
\subsection{Experimental Setup}

\textbf{Models and Hardware.}
We evaluate all methods on three video diffusion models, Wan2.2-TI2V-5B, Wan2.1-T2V-1.3B~\citep{wan2025wan}, and LTX-Video-13B~\citep{hacohen2024ltx}.
Wan2.2-TI2V-5B serves as our primary evaluation model and generates 121 frames at 1280$\times$704 resolution and 24 FPS using 50-step UniPC sampling.
Within each model, all methods use the same model-specific generation configuration, random-seed protocol, precision, and decoding pipeline.
Experiments are conducted on NVIDIA H100 GPUs.

\paragraph{Workload and Cache Settings.}
We construct the serving workload from VidProM~\citep{wang2024vidprom}, with 1,000 historical requests as a fixed warm cache and a disjoint 300-request test stream.
Within each model, all cross-request methods use the same cache and request order.
Cache misses fall back to dense generation, and all online costs, including retrieval, early probing, signature construction, compatibility filtering, reuse, fallback, and decoding, are included in end-to-end latency for cache-hit requests.

\paragraph{Baselines.}
We compare \emph{Carnator} with dense inference, NIRVANA~\citep{agarwal2024approximate}, Chorus~\citep{liu2026beyond}, and Chorus~II~\citep{liu2026chorus}.
NIRVANA is originally designed for text-to-image (T2I) generation, and we adapt its cross-request latent reuse to T2V.
Chorus is evaluated in its original T2V setting, while Chorus~II, originally developed for I2V, is adapted to T2V by transferring its historical sparse-attention reuse mechanism.
All methods share the same backbone, workload, and cache setting.
CHAI~\citep{cherian2026chai} is evaluated separately in Appendix~\ref{app:chai} due to its OpenSora~1.2-specific STDiT design.

\paragraph{Metrics.}
We evaluate both generation quality and inference efficiency.
Generation quality is measured by VBench-Q~\citep{huang2024vbench} and CLIP-T~\citep{radford2021learning}, capturing general video quality and text-condition alignment, respectively.
Efficiency is measured by end-to-end latency, DiT latency, and speedup over dense inference.

\paragraph{Candidate Recall.}
We rank historical requests by CLIP text similarity and retain the top-$K$
candidates for generation-native compatibility assessment.
We use $K=10$ for all experiments.
For the VidProM workload, noun composition is used to establish entity
correspondence between recalled requests.
We retain pairs with identical noun sets or a single noun substitution,
$
|\mathcal{N}(y_q)\setminus\mathcal{N}(y_j)|
=
|\mathcal{N}(y_j)\setminus\mathcal{N}(y_q)|
\le 1.
$

\subsection{Main Results}


Table~\ref{tab:main_results} summarizes the end-to-end performance of all methods on the VidProM workload.
On Wan2.2-TI2V-5B, \emph{Carnator} accepts fewer historical caches than every evaluated cross-request baseline (72.67\% versus 80.67\% hit rate), yet it still delivers the highest workload-level acceleration: 1.36$\times$ end-to-end speedup over dense inference against 1.24$\times$ for the strongest baseline.
This reversal is the central empirical point of our evaluation and follows directly from the argument of Fig.~\ref{fig:motivation}. Because \emph{Carnator} declines caches that are semantically similar but generation-incompatible, the computation it does accept is more extensively reusable.
Among cache-hit requests, it reaches the lowest end-to-end and DiT latencies on all three backbones, at 187.30\,s and 159.65\,s on Wan2.2-TI2V-5B (1.57$\times$ and 1.66$\times$ over dense inference), the largest per-hit savings of any evaluated method.
Normalized by hit rate, each accepted cache buys $0.50\times$ of end-to-end speedup on Wan2.2, against $0.30\times$ for the strongest baseline.
Reuse validity and reuse scope thus act in the same direction: a stricter validity rule does not merely avoid bad reuse; it makes each accepted reuse more profitable.

The efficiency gains are particularly pronounced within the DiT backbone.
Region-guided latent reuse reduces redundant token updates, while historical sparse-connectivity reuse avoids unnecessary query--key interactions during attention computation.
Their joint use therefore increases the amount of computation that can be reused from each compatible historical request.

In terms of generation quality, \emph{Carnator} achieves a VBench-Q score of 0.6382, comparable to dense inference and the evaluated acceleration baselines, together with a CLIP-T score of 0.2913.
The VBench-Q results indicate that the proposed reuse mechanism preserves the major visual-quality dimensions captured by VBench while substantially reducing inference latency.
Across the evaluated methods, the results reveal a favorable quality--efficiency trade-off for cross-request computation reuse.

Overall, the results highlight two complementary factors in cross-request acceleration.
Reuse validity determines whether historical computation can be reused, while reuse scope determines where target-specific computation must be preserved within joint latent-attention reuse.
Together, these two components enable \emph{Carnator} to translate cross-request redundancy into substantial end-to-end inference savings.

\begin{table}[t]
\centering
\caption{
End-to-end performance on the VidProM workload across three video diffusion models.
Within each model, all methods process the same request sequence under the same generation configuration.
For cross-request methods, cache misses fall back to dense generation.
}
\label{tab:main_results}
\resizebox{\textwidth}{!}{
\begin{tabular}{lcccccccc}
\toprule
Method
& CLIP-T $\uparrow$
& VBench-Q $\uparrow$
& DiT Hit Latency (s) $\downarrow$
& E2E Hit Latency (s) $\downarrow$
& DiT Speedup (Hit) $\uparrow$
& E2E Speedup (Hit) $\uparrow$
& E2E Speedup $\uparrow$
& Hit Rate \\
\midrule

\multicolumn{9}{c}{\textbf{Wan2.2-TI2V-5B}} \\
\midrule

Dense
& \textbf{0.3372} & 0.6331 & 265.23 & 293.88 & 1.00$\times$ & 1.00$\times$ & 1.00$\times$ & N/A \\

NIRVANA-VID
& 0.2955 & \textbf{0.6457} & 204.21 & 256.09 & 1.30$\times$ & 1.15$\times$ & 1.12$\times$ & 80.67\% \\

Chorus
& 0.3005 & 0.6392 & 175.48 & 264.47 & 1.51$\times$ & 1.11$\times$ & 1.09$\times$ & 80.67\% \\

Chorus II
& 0.2986 & 0.6400 & 201.58 & 222.44 & 1.32$\times$ & 1.32$\times$ & 1.24$\times$ & 80.67\% \\

\textbf{CARNATOR}
& 0.2913 & 0.6382 & \textbf{159.65} & \textbf{187.30} & \textbf{1.66$\times$} & \textbf{1.57$\times$} & \textbf{1.36$\times$} & 72.67\% \\

\midrule
\multicolumn{9}{c}{\textbf{Wan2.1-T2V-1.3B}} \\
\midrule

Dense
& \textbf{0.2984} & 0.6379 & 974.2 & 1138.6 & 1.00$\times$ & 1.00$\times$ & 1.00$\times$ & N/A \\

NIRVANA-VID
& 0.2944 & 0.6210 & 884.5 & 1035.9 & 1.10$\times$ & 1.10$\times$ & 1.08$\times$ & 80.67\% \\

Chorus
& 0.2915 & 0.6089 & 582.4 & 719.8 & 1.67$\times$ & 1.58$\times$ & 1.42$\times$ & 80.67\% \\

Chorus II
& 0.2891 & \textbf{0.6445} & 556.9 & 577.1 & 1.75$\times$ & 1.97$\times$ & 1.66$\times$ & 80.67\% \\

\textbf{CARNATOR}
& 0.2810 & 0.6361 & \textbf{496.0} & \textbf{525.58} & \textbf{1.96}$\times$ & \textbf{2.17$\times$} & \textbf{1.72$\times$} & 77.67\% \\

\midrule
\multicolumn{9}{c}{\textbf{LTX-Video-13B}} \\
\midrule

Dense
& 0.2608 & 0.5896 & 79.37 & 81.42 & 1.00$\times$ & 1.00$\times$ & 1.00$\times$ & N/A \\

NIRVANA-VID
& 0.2586 & 0.6060 & 70.78 & 72.82 & 1.12$\times$ & 1.12$\times$ & 1.09$\times$ & 80.67\% \\

Chorus
& 0.2574 & 0.5994 & 70.78 & 136.24 & 1.12$\times$ & 0.60$\times$ & 0.65$\times$ & 80.67\% \\

Chorus II
& \textbf{0.2639} & \textbf{0.6244} & 75.26 & 77.69 & 1.05$\times$ & 1.05$\times$ & 1.04$\times$ & 80.67\% \\

\textbf{CARNATOR}
& 0.2583 & 0.6022 & \textbf{62.53} & \textbf{65.88} & \textbf{1.27$\times$} & \textbf{1.24$\times$} & \textbf{1.18$\times$} & 79.33\% \\

\bottomrule
\end{tabular}
}
\end{table}

\subsection{Ablation Studies}
\label{subsec-ablation_studies}

\paragraph{Generation-Level Compatibility.}
Table~\ref{tab:compatibility_ablation} evaluates the effect of incorporating the Early Signature into cross-request cache retrieval.
Compared with text-level retrieval alone, generation-level filtering reduces the cache hit rate from 77.33\% to 72.67\%, while improving CLIP-T from 0.2725 to 0.2913 and maintaining a comparable VBench-Q score of 0.6382.
This indicates that the Early Signature filters out a small fraction of textually similar but generation-incompatible caches, leading to better text--video alignment without sacrificing overall visual quality.
Meanwhile, the end-to-end latency remains essentially unchanged, showing that the additional compatibility filtering introduces little overhead relative to the computation saved by cache reuse.

On a disjoint certification set, the risk-aware compatibility rule accepts 50 of 70 request-cache pairs with no unsafe cases.
The resulting one-sided Clopper-Pearson upper bound is 4.50 percent, satisfying the prescribed 5 percent risk constraint.
The same risk-aware threshold-selection protocol is applied across models, with details provided in Appendix~\ref{app:risk-aware}.

\begin{table}[H]
\centering
\caption{
Ablation of reuse-validity assessment on the VidProM workload.
Speedup is calculated with respect to dense generation.
}
\label{tab:compatibility_ablation}
\resizebox{\linewidth}{!}{
\begin{tabular}{lccccc}
\toprule
Retrieval / Gate
& Hit Rate
& CLIP-T $\uparrow$
& VBench-Q $\uparrow$
& E2E Hit Latency (s) $\downarrow$
& Speedup $\uparrow$ \\
\midrule

Text-Level Retrieval
& 77.33\% & 0.2725  & 0.6301 & 185.6 & 1.58$\times$ \\

+ Early Signature
& 72.67\% & 0.2913 & 0.6382 & 187.3 & 1.57$\times$ \\

\bottomrule
\end{tabular}
}
\end{table}

\paragraph{Joint Latent-Attention Reuse.}
Table~\ref{tab:reuse_ablation} evaluates the individual contributions of latent reuse and attention reuse on a randomly sampled fixed subset of 100 requests.
Latent reuse reduces redundant token updates and lowers end-to-end latency to 206.06\,s, while attention reuse decreases query--key computation and achieves 160.91\,s.
Combining the two mechanisms achieves the lowest latency of 155.85\,s and the highest end-to-end speedup of 1.71$\times$, confirming that the two forms of reuse provide complementary acceleration benefits.
Across all variants, CLIP-T and VBench-Q remain within a narrow range, indicating that the additional acceleration from joint reuse does not substantially compromise generation quality.

\begin{table}[H]
\centering
\caption{
Ablation of latent and attention reuse in \emph{CARNATOR}.
}
\label{tab:reuse_ablation}
\resizebox{\linewidth}{!}{
\begin{tabular}{lcccc}
\toprule
Variant
& Latency (s) $\downarrow$
& Speedup $\uparrow$
& CLIP-T $\uparrow$
& VBench-Q $\uparrow$ \\
\midrule

Full Computation
& 266.26 & 1.00$\times$ & 0.3391 & 0.6432 \\

Latent Reuse Only
& 206.06 & 1.29$\times$ & 0.3237 & 0.6436 \\

Attention Reuse Only
& 160.91 & 1.65$\times$ & 0.3291 & 0.6291 \\

\textbf{Joint Latent-Attention Reuse}
& 155.85 & 1.71$\times$ & 0.3281 & 0.6433 \\

\bottomrule
\end{tabular}
}
\end{table}

\section{Conclusion}

In this work, we presented \emph{Carnator}, a cross-request acceleration framework for text-to-video diffusion models that uses generation-native compatibility evidence to jointly address reuse validity and reuse scope.
\emph{Carnator} constructs an Early Signature from early diffusion states to identify reusable historical caches beyond prompt similarity, and applies risk-aware compatibility decisions.
For accepted caches, it jointly exploits redundancy in latent trajectories and attention connectivity through region-guided latent reuse and historical sparse-connectivity reuse.
Experiments demonstrate that \emph{Carnator} reduces video diffusion inference latency while maintaining competitive generation quality across diverse generation requests.
More broadly, our results show that cross-request redundancy provides a complementary dimension for accelerating video generation beyond conventional optimization within individual denoising trajectories.
\section*{AI Use Statement}

Generative AI tools were used to assist with language editing and manuscript organization.
They were also used to help refine figures, captions, and terminology. 
All AI-assisted content was reviewed and verified by the authors.

\newpage

\bibliography{iclr2027_conference}
\bibliographystyle{iclr2027_conference}

\appendix
\section{Additional Analysis}
\label{app:additional_analysis}

\subsection{Early-State Generation-Native Compatibility Evidence}

\begin{figure}[H]
    \centering
    \includegraphics[width=\textwidth]{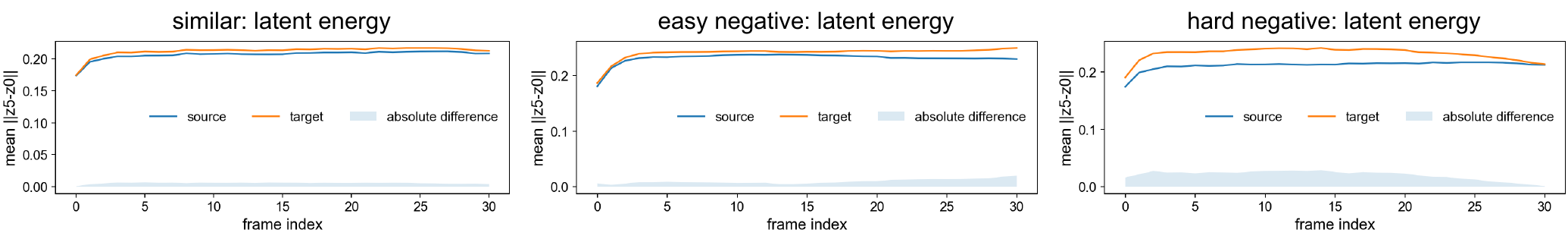}
    \caption{
    Latent-state discrepancy for similar, easy-negative, and hard-negative request pairs across video frames.
    }
    \label{fig:early_signature_energy}
\end{figure}

Fig.~\ref{fig:early_signature_energy} provides additional analysis of early latent dynamics.
Similar request pairs exhibit consistently smaller source--target discrepancies than incompatible pairs, providing further evidence that early diffusion states contain useful generation-native compatibility evidence for reuse-validity assessment.

\subsection{Cross-Request Attention Similarity}

\begin{figure}[H]
    \centering
    \includegraphics[width=\textwidth]{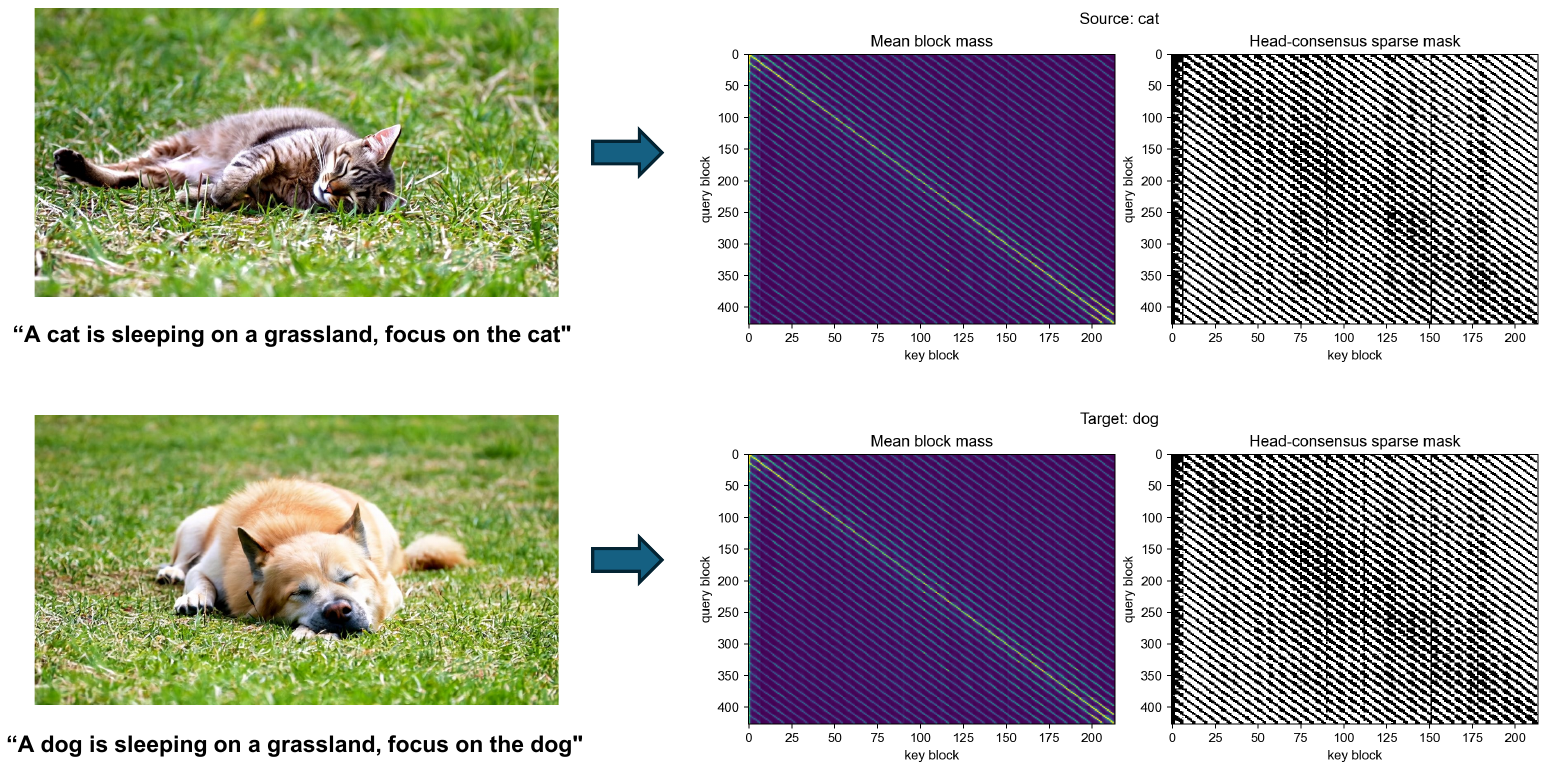}
    \caption{
    Blockwise attention mass and sparse connectivity for two related text-to-video requests.
    Similar requests preserve substantial structural similarity in their attention patterns despite changes in object identity.
    }
    \label{fig:attention_similarity}
\end{figure}

Fig.~\ref{fig:attention_similarity} visualizes the blockwise attention structure of two semantically related requests.
Despite the object substitution, the two requests exhibit highly similar attention mass distributions and sparse connectivity patterns, motivating Historical Sparse-Connectivity Reuse.

\section{Additional Qualitative Results}
\label{app:qualitative}

This section provides additional qualitative results of \emph{Carnator} on diverse text-to-video generation requests.
The examples illustrate that \emph{Carnator} preserves target-specific content and overall visual quality while reusing computation from compatible historical requests.

\begin{figure}[H]
    \centering
    \includegraphics[width=\linewidth]{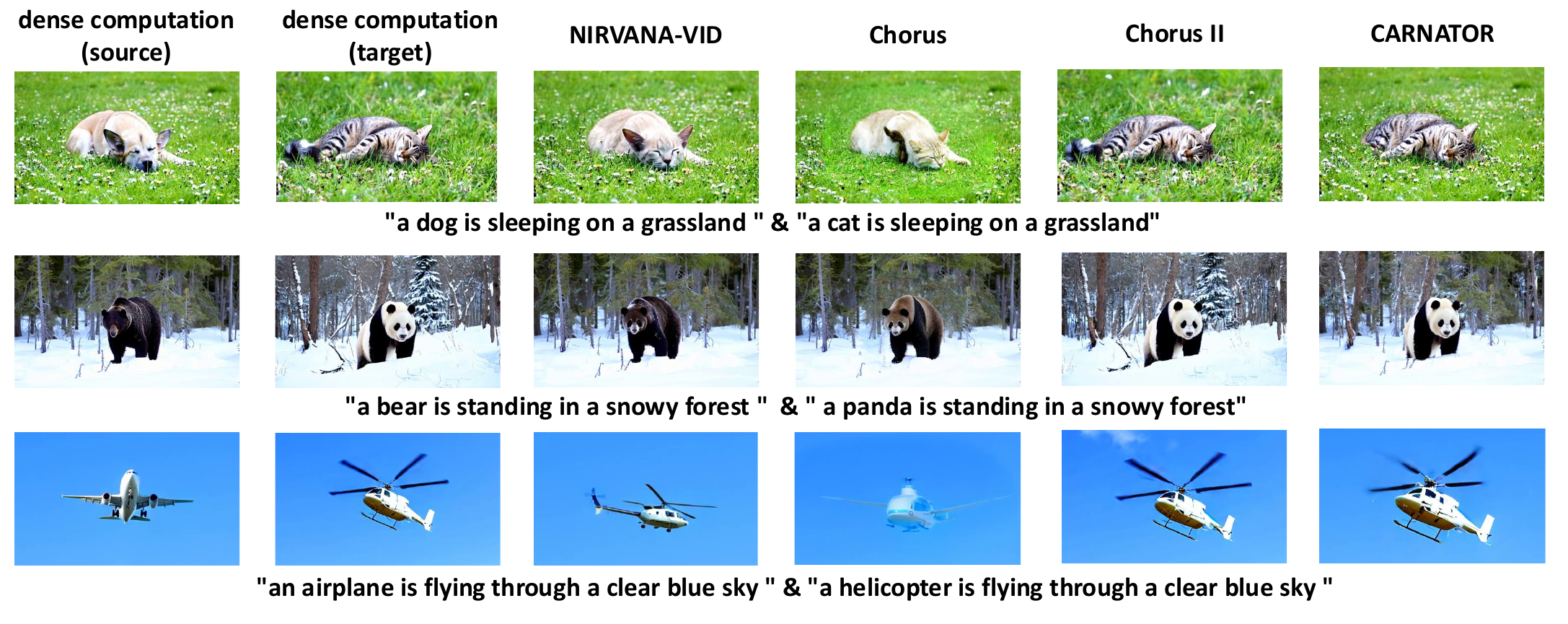}
    \caption{
    Qualitative result of \emph{Carnator}
    }
    \label{fig:chai_portability}
\end{figure}
\section{Early Signature Details}
\label{app:early-signature}

\subsection{Evidence Region Construction}
We construct the Early Signature after the first four denoising steps.
Noun entities are extracted using spaCy with the \texttt{en\_core\_web\_sm}
model, retaining tokens tagged as nouns.
For each noun, we collect its cross-attention response from a fixed
intermediate Transformer layer and aggregate the response across attention
heads and the corresponding text-token positions.
High- and low-response visual tokens are selected as noun and background
seeds, respectively. These seeds are propagated through self-attention to
obtain a noun--background contrast score. The cross- and self-attention
responses are fused as
\[
F = \operatorname{Norm}
\left[
S\left(\epsilon + (1-\epsilon)C^{\gamma}\right)
\right].
\]
Fixed fusion coefficients and hysteresis thresholds are used to obtain the
Evidence Region $\mathcal{R}_{\mathrm{evidence}}$, and the same configuration
is used throughout evaluation for each model.

\paragraph{Signature descriptors.}
For each noun, the implementation records the Evidence Region together with
its normalized support size---computed on the tail slice $m_{r,o}(T{-}1,\cdot,\cdot)$ and normalized by $HW$---and appearance, while object shape is derived from that same tail slice during comparison.
Appearance is extracted by decoding the latent once and projecting the
Evidence Region onto the final decoded frame.
Pixels inside the region are converted from sRGB to CIELAB under the D65
illuminant and averaged to form the appearance descriptor.
For compatibility comparison, shared nouns are compared using CIEDE2000
color distance, while a one-noun substitution additionally uses log-area
ratio, normalized mask-shape distance, and CIEDE2000 appearance distance.
Noun strings are normalized to lowercase, repeated nouns are merged, and the
current implementation accepts either identical noun sets or a single
one-to-one noun substitution.

\subsection{Early Probe Evidence Region}
\label{app:early_mask}

\begin{figure}[H]
    \centering
    \includegraphics[width=\textwidth]{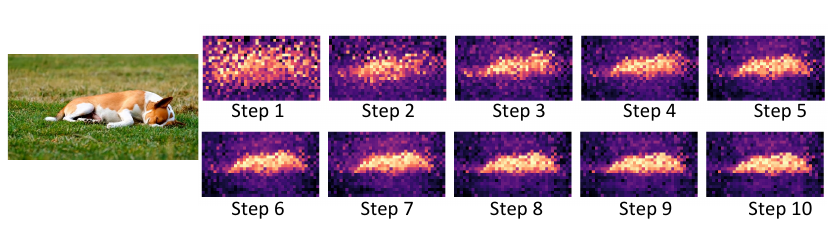}
    \caption{
    Heat Map of Evidence Region $\mathcal{R}_{\mathrm{evidence}}$ under the first ten denoising steps.
    }
    \label{fig:early_evidence_region}
\end{figure}

Figure~\ref{fig:early_evidence_region} shows that the Evidence Region rapidly converges during the first few denoising steps.
The response is diffuse at the beginning, rapidly concentrates on the target object within the first few steps, and becomes spatially stable thereafter.
This observation provides qualitative evidence that useful localization cues emerge early in the denoising trajectory, allowing \emph{Carnator} to construct its Early Signature from a lightweight early probe rather than requiring a complete generation.

\subsection{Reuse-Scope Implementation Details}
\label{app:reuse-scope}

Source trajectories are generated with the same model-specific scheduler and
denoising schedule used for target inference, and post-scheduler latents are
stored by timestep. For reuse, the target resumes from the corresponding
historical latent at the reuse point and thereafter accesses the cached latent
from the same scheduler step, while target prompt conditioning and guidance
computations are recomputed for the target request.
The cached Evidence Region is spatially dilated to form the Active and Context
Regions, with the Context Region using the larger support.
Region-restricted updates are applied to the designated middle Transformer
layers, while the remaining layers follow the historical sparse-connectivity
path.
We obtain the Active Region by dilating the Evidence Region with a radius of 2 tokens, and the Context Region by a further dilation with a radius of 3 tokens, applied per latent frame on the $H\times W$ grid.
The band
$\mathcal{R}_{\mathrm{active}}
\setminus
\mathcal{R}_{\mathrm{evidence}}$
is used as a transition region for latent fusion.
Within this band, the source weight is the Chebyshev distance to the Evidence Region, normalized by the maximum distance attained in the band and mapped through a smoothstep function, yielding a gradual transition from target to historical latent states.

Historical self-attention is partitioned into query--key blocks, and for each
head and query block we retain the smallest set of key blocks covering a fixed
fraction of the source attention mass.
The resulting binary connectivity masks are stored per denoising step,
guidance branch, and Transformer layer.
During target inference, $Q$, $K$, and $V$ are recomputed from the target
generation, while only the historical block connectivity is reused.
Within the middle layers, the cached connectivity is further restricted to
active queries and their valid contextual tokens before sparse execution.

\subsection{Overall Inference Procedure}
\label{app:overall-procedure}

Algorithm~\ref{alg:carnator} summarizes the end-to-end inference procedure of
\emph{Carnator}. The risk-aware compatibility thresholds are selected offline
as described in Appendix~F and remain fixed during serving. For each target
request, \emph{Carnator} first retrieves candidate histories, constructs the
target Early Signature, and selects the first compatible cache in recall order.
If no compatible cache is found, the request falls back to dense generation.
Otherwise, the accepted cache is used for joint latent-attention reuse.

\begin{algorithm}[H]
\caption{\textsc{Carnator} Cross-Request Inference}
\label{alg:carnator}
\begin{algorithmic}[1]
\Require Target request $q$, cache bank $\mathcal{B}$,
         recall size $K$, compatibility thresholds $\boldsymbol{\tau}^{\star}$
\Ensure Generated video $V_q$

\State $\mathcal{L}_q \gets \textsc{TextRecall}(q,\mathcal{B},K)$
\State $(S_q,\mathcal{R}_{\mathrm{evidence}})
       \gets \textsc{EarlyProbe}(q)$
\State $\mathcal{C}^{\star} \gets \varnothing$

\For{$\mathcal{C}_j \in \mathcal{L}_q$ in recall order}
    \State $\mathbf{d}_j \gets
           \textsc{Compatibility}(S_q,S_j)$
    \If{$g_{\boldsymbol{\tau}^{\star}}(q,\mathcal{C}_j)=1$}
        \State $\mathcal{C}^{\star} \gets \mathcal{C}_j$
        \State \textbf{break}
    \EndIf
\EndFor

\If{$\mathcal{C}^{\star}=\varnothing$}
    \State \Return $\textsc{DenseGeneration}(q)$
\EndIf

\State $(\mathcal{R}_{\mathrm{active}},
        \mathcal{R}_{\mathrm{context}},\mathbf{W})
        \gets
        \textsc{BuildReuseRegions}(\mathcal{R}_{\mathrm{evidence}})$

\State Initialize target generation from the accepted cached trajectory

\For{each remaining denoising step $k$}
    \State Recompute target-specific latent states in
           $\mathcal{R}_{\mathrm{active}}$
    \State Fuse target and historical latents using $\mathbf{W}$
    \State Recompute target $Q$, $K$, and $V$
    \State Reuse historical sparse connectivity within the valid
           target computation domain
\EndFor

\State $V_q \gets \textsc{Decode}(\mathbf{z}_q)$
\State \Return $V_q$

\end{algorithmic}
\end{algorithm}

\section{Cross-Architecture Portability of CHAI}
\label{app:chai}

\begin{figure}[H]
    \centering
    \includegraphics[width=\linewidth]{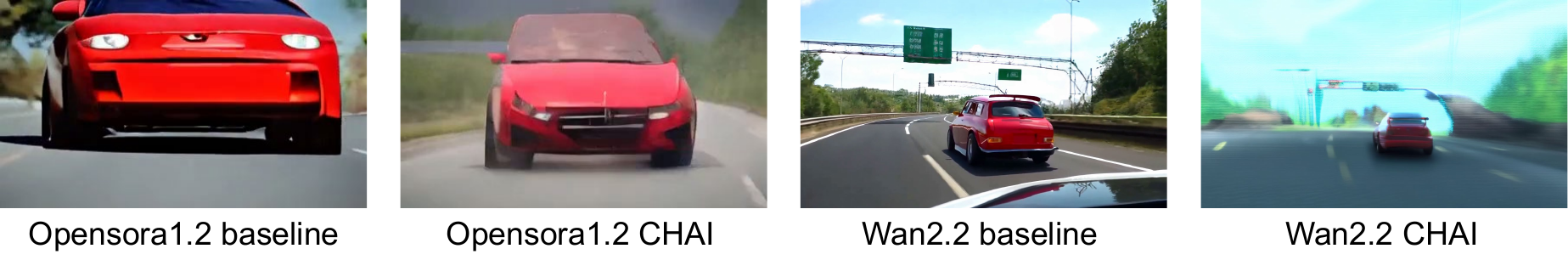}
    \caption{
    Qualitative comparison of CHAI across video diffusion architectures.
    From left to right, we show the OpenSora 1.2 baseline, CHAI on OpenSora 1.2, the Wan2.2 baseline, and a direct adaptation of CHAI to Wan2.2.
    CHAI preserves generation quality on its native OpenSora architecture, while the direct Wan2.2 adaptation exhibits noticeable visual degradation.
    }
    \label{fig:chai_comparison}
\end{figure}

CHAI~\citep{cherian2026chai} designs Cache Attention specifically for the STDiT architecture of OpenSora 1.2.
We further examine its cross-architecture portability by directly adapting the same cache-reuse mechanism to Wan2.2.
As illustrated in Fig.~\ref{fig:chai_comparison}, CHAI preserves generation quality on its native OpenSora 1.2 architecture, whereas the direct Wan2.2 adaptation exhibits noticeable visual degradation.
This qualitative comparison suggests that CHAI's Cache Attention mechanism requires architecture-specific adaptation when transferred to models other than OpenSora~1.2.

\section{Layer-Window Ablation}
\label{sec:app-layer-window}

We ablate the number of middle Transformer layers used for region-restricted
latent updates on Wan2.2-TI2V-5B. All configurations use the same accepted
request--cache pair, while the remaining layers perform historical
sparse-connectivity reuse.

\begin{table}[H]
\centering
\small
\setlength{\tabcolsep}{4.5pt}
\caption{\textbf{Layer-window ablation on Wan2.2-TI2V-5B.}
All reuse variants use the same accepted cache.
Latency includes the dense probe, and speedup is relative to dense inference.}
\label{tab:app-layer-window}
\begin{tabular}{l c c c c c}
\toprule
Region-update layers
& Window
& Latency (s)$\downarrow$
& Speedup$\uparrow$
& CLIP-T$\uparrow$
& VBench-Q$\uparrow$ \\
\midrule
None (dense)      & ---      & 273.61 & 1.00$\times$ & 0.3330 & 0.6699 \\
\midrule
5                 & 13--17   & 160.67 & 1.70$\times$ & 0.2681 & 0.6447 \\
11                & 10--20   & 156.67 & 1.75$\times$ & 0.2676 & 0.6415 \\
15                & 8--22    & 150.58 & 1.82$\times$ & 0.2700 & 0.6440 \\
20                & 6--25    & 150.97 & 1.81$\times$ & 0.2803 & 0.6497 \\
25                & 3--27    & 148.97 & 1.84$\times$ & 0.2749 & 0.6268 \\
30                & 0--29    & 145.09 & 1.89$\times$ & 0.2617 & 0.6092 \\
\bottomrule
\end{tabular}
\end{table}

As shown in Table~\ref{tab:app-layer-window}, wider region-update windows provide
modest additional speedup, while generation quality begins to degrade beyond
20 layers.

\section{Risk-Aware Compatibility Decision}
\label{app:risk-aware}

For each request--cache pair, we generate a dense reference under the same
target prompt and evaluate the reused and dense generations under the same
quality criteria.
Following Sec.~\ref{subsec-cache_compatibility}, a reuse outcome is considered
acceptable when
\[
y_i =
\mathbb{I}
\left[
Q_i^{(m)}
\ge
Q_i^{\mathrm{dense},(m)}-\Delta_m,
\ \forall m\in\mathcal{M}
\right].
\]
In our evaluation, we instantiate
$\mathcal{M}=\{\mathrm{CLIP\text{-}T},\mathrm{VBench\text{-}Q}\}$
and set
$\Delta_{\mathrm{CLIP\text{-}T}}
=
\Delta_{\mathrm{VBench\text{-}Q}}
=
0.02$.
Dense references are used only offline to construct reuse-outcome labels for
threshold selection and certification, and are not required during serving.

We use a threshold-selection set of 30 request--cache pairs and a disjoint
certification set of 70 pairs.
On the threshold-selection set, \emph{Carnator} selects the compatibility
thresholds to maximize reuse coverage subject to the risk constraint in
Sec.~\ref{subsec-cache_compatibility}, yielding
\[
\boldsymbol{\tau}^{\star}
=
(\tau_{\mathrm{context}},
 \tau_{\mathrm{size}},
 \tau_{\mathrm{shape}},
 \tau_{\mathrm{color}})
=
(1.62, 2.17, 1.00, 7.58).
\]

We set the target unsafe-reuse rate to $\delta=5\%$ and use a one-sided
confidence level of $90\%$ ($\alpha=0.10$).
On the certification set, the selected risk-aware compatibility rule accepts
50 of the 70 request--cache pairs, with no unsafe cases.
The corresponding reuse coverage is $71.43\%$, and the one-sided
Clopper--Pearson upper bound is
$U_{0.10}(0,50)=4.50\%$, satisfying the prescribed risk constraint.

\begin{table}[H]
\centering
\small
\caption{
Risk-aware threshold selection and certification results.
The target unsafe-reuse rate is $\delta=5\%$ at a one-sided confidence level
of $90\%$.
}
\label{tab:risk-aware}
\begin{tabular}{c c c c c c}
\toprule
$\boldsymbol{\tau}^{\star}$
& Coverage
& Accepted $n$
& Unsafe $u$
& Empirical Risk
& $U_{\alpha}(u,n)$ \\
\midrule
$(1.62,\,2.17,\,1.00,\,7.58)$
& 71.43\%
& 50
& 0
& 0.00\%
& 4.50\% \\
\bottomrule
\end{tabular}
\end{table}
\section{Latency Accounting}
\subsection{End-to-end latency}
E2E latency is the wall-clock time of one request: the timer opens at the start of the request and closes when the decoded video is ready. It therefore covers retrieval, text encoding and initial latent construction, the entire denoising loop (DiT forwards and scheduler updates, latent injection/fusion, and in-loop mask or index lookups), any dense probe prefix a method requires, method preprocessing that is not hoisted out (e.g. Chorus's LLM prompt-diff and segmentation), and VAE decoding of the final latent.

\subsection{DiT latency}
DiT latency isolates the denoiser: the CUDA-synchronized wall time of the transformer forward passes alone, summed over the executed steps. Synchronizing at both ends means the interval contains only that forward's own kernels — it never absorbs the launch tail of the preceding operation, nor the scheduler update, CFG/STG combination, latent injection or fusion, mask lookups, VAE decoding, or text encoding. Under classifier-free guidance the model runs once per branch and all branches are counted, so the metric scales with the number of forwards rather than sampler steps; where a pipeline batches its guidance branches into a single call, that call counts once. Because a method may add work before reuse, any dense probe prefix is included here too, and its steps are added to the executed-step count, so the reported DiT time always covers every forward the request actually ran. A per-module CUDA-event profiler (self-attention, cross-attention, FFN) is recorded alongside as a cross-check.



\end{document}